\documentclass[10pt,twocolumn,letterpaper]{article}

\usepackage{cvpr}              % To produce the CAMERA-READY version
\usepackage[cmintegrals]{newtxmath}
\usepackage{graphicx}
\usepackage{algorithm}
\usepackage{algorithmic}
\usepackage{booktabs}

\usepackage{multirow}
\usepackage{subfloat}
\usepackage{bbding}
\usepackage{xcolor}
\usepackage[pagebackref,breaklinks,colorlinks]{hyperref}

\usepackage[capitalize]{cleveref}
\crefname{section}{Sec.}{Secs.}
\Crefname{section}{Section}{Sections}
\Crefname{table}{Table}{Tables}
\crefname{table}{Tab.}{Tabs.}

\def\cvprPaperID{3400} % *** Enter the CVPR Paper ID here
\def\confName{CVPR}
\def\confYear{2023}

\begin{document}
	
\title{Context-aware Mixture-of-Experts for Unbiased Scene Graph Generation}

\author{Liguang Zhou\textsuperscript{1}, Yuhongze Zhou\textsuperscript{2}, Tin Lun Lam\textsuperscript{1}\thanks{Corresponding Author.} , Yangsheng Xu\textsuperscript{1}\\
	\textsuperscript{1} AIRS, CUHK-Shenzhen\\
	\textsuperscript{2} McGill University, Montreal, Canada \\
	{\tt\small \{liguangzhou@link., tllam@, ysxu@\}cuhk.edu.cn,}
	{\tt\small yuhongze.zhou@mcgill.ca}
	% To sav
}

\maketitle

%%%%%%%%% ABSTRACT
\begin{abstract}
	Scene graph generation (SGG) has gained tremendous progress in recent years. However, its underlying long-tailed distribution of predicate classes is a challenging problem. For extremely unbalanced predicate distributions, existing approaches usually construct complicated context encoders to extract the intrinsic relevance of scene context to predicates and complex networks to improve the learning ability of network models for highly imbalanced predicate distributions. To address the unbiased SGG problem, we introduce a simple yet effective method dubbed Context-Aware Mixture-of-Experts (CAME) to improve model diversity and mitigate biased SGG without complicated design. Specifically, we propose to integrate the mixture of experts with a divide and ensemble strategy to remedy the severely long-tailed distribution of predicate classes, which is applicable to the majority of unbiased scene graph generators. The biased SGG is thereby reduced, and the model tends to anticipate more evenly distributed predicate predictions. To differentiate between various predicate distribution levels, experts with the same weights are not sufficiently diverse. In order to enable the network dynamically exploit the rich scene context and further boost the diversity of model, we simply use the built-in module to create a context encoder. The importance of each expert to scene context and each predicate to each expert is dynamically associated with expert weighting (EW) and predicate weighting (PW) strategy. We have conducted extensive experiments on three tasks using the Visual Genome dataset, showing that CAME outperforms recent methods and achieves state-of-the-art performance. Our code will be available publicly.
\end{abstract}

%With the help of the mixture of experts, a divide and ensemble strategy  is used to handle the long-tailed distribution of predicates.
% Almost all existing scene graph generation (SGG) approaches follow the same framework, using a similar backbone network for object detection and a customized network for scene graph generation.

%%%%%%%%% BODY TEXT
\section{Introduction}
\label{sec:intro}

\begin{figure}[t]
	\centering
	\subfloat{
		\begin{minipage}{0.5\textwidth}
			\centering
			\includegraphics[width=5cm,height=5cm]{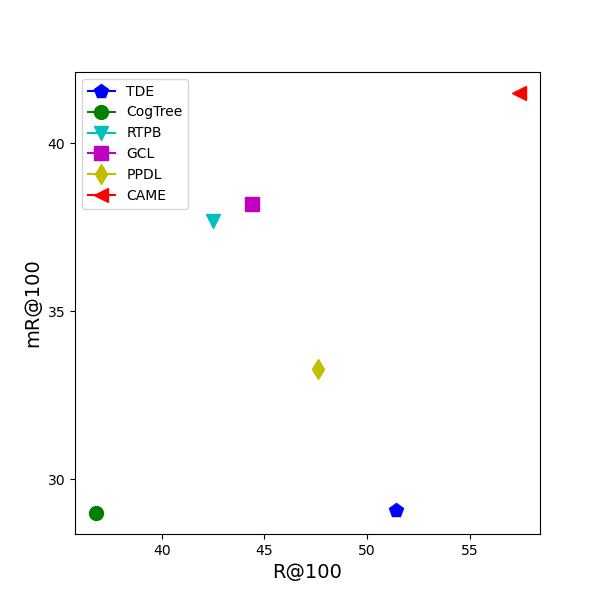} 
	\end{minipage}}
	\caption{The figure compares CAME and current state-of-the-art USGG methods on the mR@100 and R@100 of the PredCls task. The results show CAME has significantly improved on both  mR@100 and R@100, reaching an unbiased SGG performance.}
	%\caption{Differences between our method and conventional unbiased SGG methods. a) Conventional methods only train a single model on the Visual Genome dataset, which is inferior in dealing with the intrinsic heavily biased relationship distributions b) Our method leverages the mixtures of experts to deal with the long-tailed distribution in a divide and ensemble manner. Meanwhile, the context information is further utilized for building the context-aware mixture-of-experts (CAME), which increases the diversity of the model. }
	\label{fig:usgg_case}
\end{figure}

%\begin{figure}[t]
%	\centering
%	\subfloat{
%		\begin{minipage}{0.5\textwidth}
%			\centering
%			\includegraphics[width=8cm]{Figs//Fig_1.pdf}
%	\end{minipage}}
%	\caption{Differences between our method and conventional unbiased SGG methods. a) Conventional methods only train a single model on the Visual Genome dataset, which is inferior in dealing with the intrinsic heavily biased relationship distributions b) Our method leverages the mixtures of experts to deal with the long-tailed distribution in a divide and ensemble manner. Meanwhile, the context information is further utilized for building the context-aware mixture-of-experts (CAME), which increases the diversity of the model. }
%	\label{fig:usgg_case}
%\end{figure}

\begin{figure}[htbp]
	\begin{center}
		%\resizebox{\columnwidth}{!}{
		\begin{tabular}{ccc}
			\includegraphics[width=2.5cm]{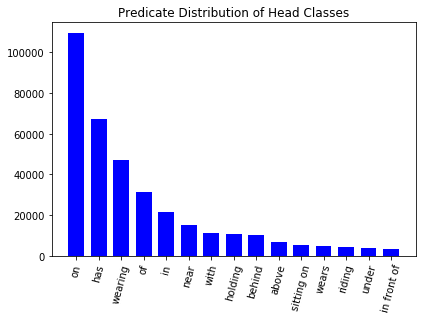} &
			\includegraphics[width=2.5cm]{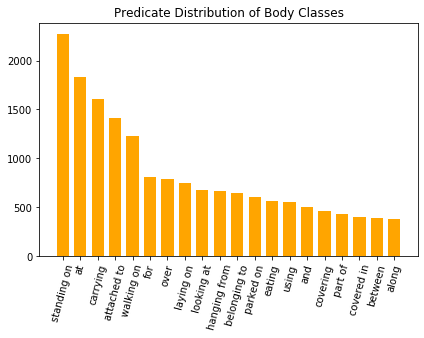} & 
			\includegraphics[width=2.5cm]{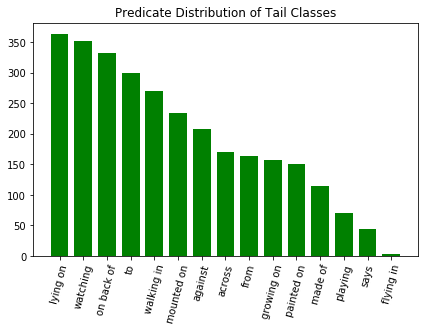}  \\
			%			a) {mean:23456,\\median:10764} &
			%			b) {mean:848,\\ median:663} &
			%			c) {mean:195, \\median:171} \\
		\end{tabular}
		\caption{Predicate distributions of \textcolor{blue}{head} classes,  \textcolor{orange}{body} classes and \textcolor{green}{tail} classes of Visual Genome dataset. The mean and median of the dataset are displayed.} 
		
		\label{fig:data_distribution} 
	\end{center}
\end{figure}

SGG has recently made notable progress, thanks to the advanced object detection and image classification algorithms \cite{xu2017scene,li2017scene,yang2018graph,gu2019scene,lin2020gps}. These techniques adhere to the standard SGG paradigm. To infer the objects in the image, object detection is applied first. Then, the retrieved object features are applied to predicate prediction using the standard cross-entropy loss. The distribution of predicate classes is severely imbalanced as shown in Fig.~\ref{fig:data_distribution}. The biased visual relationship distributions in this process, however, lead to biased predicate prediction, which ignores uncommon predicates in favor of common predicates.

%The intrinsic distribution of predicates of the Visual Genome dataset is shown in Fig.~\ref{fig:data_distribution}. We specifically displayed the predicate class distribution of the Visual Genome dataset, including three parts: head classes, body classes, and tail classes. The distribution of predicate classes is severely imbalanced. For instance, the mean values of the three parts are 23456, 848, and 195, respectively. As a result, dealing with complete highly imbalanced head and body classes, as well as tail classes with small sample sizes, presents challenges in the unbiased SGG.

Various strategies have been developed to deal with this heavy long-tailed problem. For instance, the Total Difference Effect is leveraged in the inference phase based on the causality information \cite{tang2020unbiased}. Additionally, the semantic information of the scene is well-exploited for unbiased SGG \cite{khandelwal2021segmentation}. Unlike these methods, the cognition tree (CogTree) \cite{yu2020cogtree} presents a hierarchical cognitive structure from biased predictions. Under the guidance of such a coarse-to-fine procedure, the tail classes receive more attention. Moreover, energy loss function with GAN attached to SGG models is also proposed to enhance the learning ability of the models \cite{suhail2021energy}. Such SGG methods could, however, run into the following issues: 1) there are different levels of predicate distributions on the Visual Genome dataset, but they only learn a common feature space for multiple predicate distributions, which could be very limited for realizing the unbiased SGG (USGG). Recently, the mixture of experts (ME) \cite{jacobs1991adaptive} has demonstrated effectivity in increasing the learning capacity of the model by mixing multiple networks with complementary information like the different distributions of dataset \cite{wang2019learning, dai2021generalizable,riquelme2021scaling}.
Inspired by this, we propose to use the mixture of experts to increase the learning capabilities for heavily long-tailed predicate classes, which is naturally suitable for dealing with such problems. 
2) the SGG problem has an intrinsic correlation between contextual information and predicate classes.   
This correlation has not been well exploited by mixture of experts. To exploit the intrinsic correlation between context and predicates, we propose a context-aware encoder with a built-in module that relates the intrinsic relationship between contextual information and potential predicates, thus improving the cognitive power of the model.

To address the two issues mentioned above, we propose a new approach, called Context-Aware Mixture-of-Experts (CAME), for USGG, as shown in Fig.~\ref{fig:method}. Based on a shared layer with a common encoder and decoder network, our CAME constructs a set of predicate classifiers that greatly extend the learning capability of the baseline SGG approaches. Besides, the knowledge of different experts is ensembled as the complementary information for model training, which is well suited for imbalanced predicate distributions and associates experts with specific contextual information. We believe that instead of using a single model to learn heavily imbalanced long-tailed data, we can train multiple experts to address the long-tailed distribution dataset in a divide and ensemble manner. Hence, each expert could learn complementary information guided by the designed context-aware loss. As a result, their ensemble can  represent the long-tailed predicate distribution well, thus alleviating the long-tailed problem. Specifically, mixture of experts can improve not only the discriminability of the model, but also the diversity of the model, thus reducing the biased predictions over the entire classes and leading to an unbiased SGG algorithm. However, in traditional mixture of experts models, each expert has the same importance for the input context, which reduces the model diversity and ignores the intrinsic relationship between contextual information and predicates. Therefore, we present a context-aware encoder network to dynamically measure the importance of each expert with the proposed expert weighting (EW) module. In addition, the importance of each predicate to each expert is dynamically assigned with the proposed predicate weighting (PW) module.

Based on the importance measurement, we can adaptively combine the feature of each expert into one powerful expert. With the context-aware encoder network, given the rich contextual information, higher weights will be assigned to more relevant experts. Hence, more relevant experts will provide more effective information to enhance unbiased prediction and increase the diversity of the model.

In Fig.~\ref{fig:usgg_case}, we compare our CAME method with the current state-of-the-art methods for the PredCls task of R@100 and mR@100 on the Motifs baseline. The current techniques, including TDE, CogTree, RTPB, GCL, and PPDL, demonstrate good improvements in USGG. However, as mR@100 improves over time, R@100 decreases sharply. This phenomenon shows an unbalanced improvement, i.e., in improving  the tail classes, the algorithm shifts its attention from the head to the tail, leading to another biased prediction. In contrast, CAME not only retains the highest R@100, but also gains a tremendous improvement in mR@100, showing a balanced trade-off between head, body, and tail classes through a mixture of experts and context-aware encoder network.

%In Figure~\ref{fig:usgg_case}, we compare our CAME method with the conventional method through a \textbf{subject-predicate-object} triplet example. The conventional method predicts \textbf{person-on-surfboard} while our CAME predicts \textbf{person-lying on-surfboard}. As observed from Figure~\ref{fig:data_distribution}, the \textbf{on} belongs to head classes while the \textbf{lying on} belongs to tail classes. It is clear that the conventional method tends to make biased predictions as \textbf{on} in the head classes. However, our proposed CAME can make correct and unbiased predictions as \textbf{lying on} in tail classes. Moreover, the proposed CAME method is general; hence it can be integrated to train most existing SGG frameworks. We conduct extensive experiments on the Visual Genome dataset to validate the effectiveness of our CAME method.

To summarize, to the best of our knowledge, we propose the first, context-aware mixture of experts model for unbiased SGG. It can be integrated into most existing scene graph generation networks to enhance their performance. The proposed CAME framework increases the diversity of the model because each expert can learn complementary knowledge and our model aggregates the knowledge of different experts. In addition, a context-aware encoder network is tailored for long-tailed SGG cases. Object proposals, object labels, and relations between each object pair are well represented to dynamically assign the importance of each expert and the relevance of each predicate to each expert, thus further increasing diversity. Furthermore, the context-aware loss is utilized for mitigating heavily long-tailed distribution and increasing model's diversity.

\section{Related Work}

\subsection{Unbiased Scene Graph Generation}
Scene graph generation has gained remarkable progress and many works have emerged in recent years \cite{yang2018graph,zhao2021semantically,qi2019attentive,wang2019exploring,zhang2019graphical,xu2020scene,hung2020contextual,chen2019knowledge,gu2019scene,zareian2020bridging,zareian2020learning,wang2020sketching,zhang2022boosting,cong2022reltr}. The development of SGG opened the door for future applications such as image captioning \cite{gu2019unpaired,yang2019auto,zhong2020comprehensive}, robotic applications \cite{armeni20193d,rosinol20203d, kenfack2020robotvqa,rosinol2021kimera,zhu2021hierarchical}, image manipulation \cite{dhamo2020semantic}, and cross-media retrieval \cite{peng2019unsupervised}. The neural motif is proposed and demonstrates the benefits of an LSTM-based network by using the message passing mechanism \cite{zellers2018neural}. VCTree shows the advantages of the tree structure for scene graph generation \cite{tang2019learning}. In terms of framework and open-sourced materials, they have both laid a strong basis for SGG in framework and open-sourced resources \cite{zellers2018neural, tang2019learning,tang2020sggcode}. The long-tailed predicate distribution, however, remains a challenging topic that has  attracted a lot of attention \cite{tang2020unbiased,yu2020cogtree,suhail2021energy,khandelwal2021segmentation,li2021bipartite,lu2021context,chiou2021recovering,guo2021general}.

To alleviate the biased SGG process, various approaches were developed. These methods are mainly categorized into four classes. The first type of method relies on the prior information of datasets such as the causality relations, predicate correlation, segmentation maps, probabilistic uncertainty modeling, label frequencies, predicates data distributions, resistance training and fine-grained predicates learning \cite{tang2020unbiased, zhou2022debiased, yan2020pcpl, khandelwal2021segmentation,lu2021context, chiou2021recovering,yang2021probabilistic, guo2021general, chen2022resistance}. The second type of method resorts to the optimized message passing mechanisms and efficient data representations, such as graph property sensing network, bipartite graph network, sequence to sequence transformers \cite{lin2020gps,li2021bipartite,lu2021context}. The third type of method seeks to construct an unbiased loss function to alleviate the biased learning process of original networks \cite{yu2020cogtree, suhail2021energy,li2022ppdl}, including CogTree loss, predicate probability distribution based Loss, energy based loss.

The last type of method uses multiple branches to solve the USGG problem in a divide-and-conquer manner. For instance, group collaborative learning (GCL) is proposed to treat the USGG as an incremental learning problem by constructing sets of FC layers to incrementally learn to deal with the USGG from head to tail classes \cite{dong2022stacked}. The divide and conquer method is presented with a major classifier to learn the coarse relation representation knowledge and several classifiers to learn the fine relation representation \cite{han2022divide}. They all use sets of classifiers and knowledge distillation to learn the knowledge among these classifiers. However, they do not consider enhancing the learning ability of the network itself between scene context and predicate classes.

Unlike these methods, we present a novel and general framework to enhance the learning ability of the network model. Inspired by the structure of mixture-of-experts, we propose a novel Context-Aware Mixture-of-Experts (CAME), which is tailored to achieve unbiased SGG. Concretely, our CAME incorporates multiple experts, to tackle the heavily long-tailed relation predicates distribution in a divide and ensemble manner. The bias can be well-exploited by multiple experts and then the model can be more generalized to tail classes. In addition, scene context information such as object proposals, object labels, and object relationships is used to identify the relevance between scene context and its inherent predicates. This further increases the model's diversity.

\subsection{Mixture of Experts}
The Mixture of experts is originally designed with a set of expert networks composed of an expert layer where each expert learns a subtask for the whole dataset \cite{jacobs1991adaptive}. It has been proposed for decades and successfully applied in various tasks for network enhancement, including image recognition \cite{wang2021longtailed,riquelme2021scaling}, video captioning \cite{wang2019learning}, and person Re-ID \cite{dai2021generalizable}. The traditional mixture of experts usually has the same weights for each expert or has different gating schemes to selectively turn off part of the experts in the network. For instance, the divide and co-training strategy are proposed to enhance the learning ability of the original network \cite{zhao2020towards}. The weight of each expert is contributed equally to the whole network during the learning process, which limits the network to learn more diversified knowledge.

Different from the traditional mixture of experts, we first introduce the Context-Aware Mixture-of-Experts (CAME) for unbiased SGG. To encourage the diversity of mixture-of-experts and construct the relation between the scene context and predicates, the proposed CAME is designed with EW and PW modules to assign the importance of each expert based on context information and relevance of each predicate to each expert during the learning process.

\begin{figure}[t]
	\centering
	\subfloat{
		\begin{minipage}{0.5\textwidth}
			\centering
			\includegraphics[width=\linewidth]{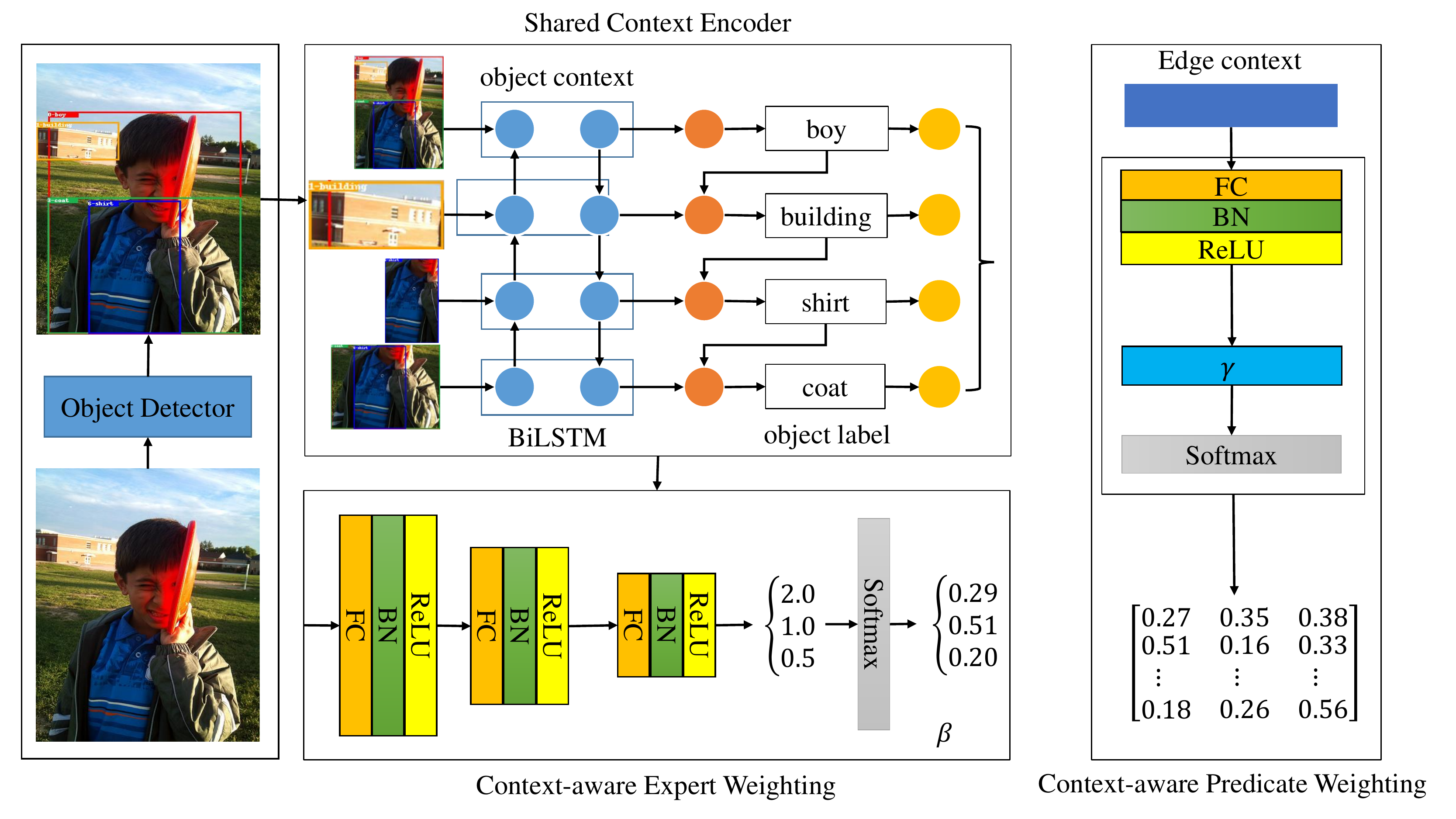}
	\end{minipage}}
	\caption{The framework of proposed context-aware encoder module. The shared context encoder is borrowed from the baseline network for efficiency and simplicity.}
	\label{fig:ca_ew_pw}
\end{figure}

\section{Approach}
\begin{figure*}[t]
	\centering
	\subfloat{
		\begin{minipage}{1\textwidth}
			\centering
			\includegraphics[width=15cm,height=7cm]{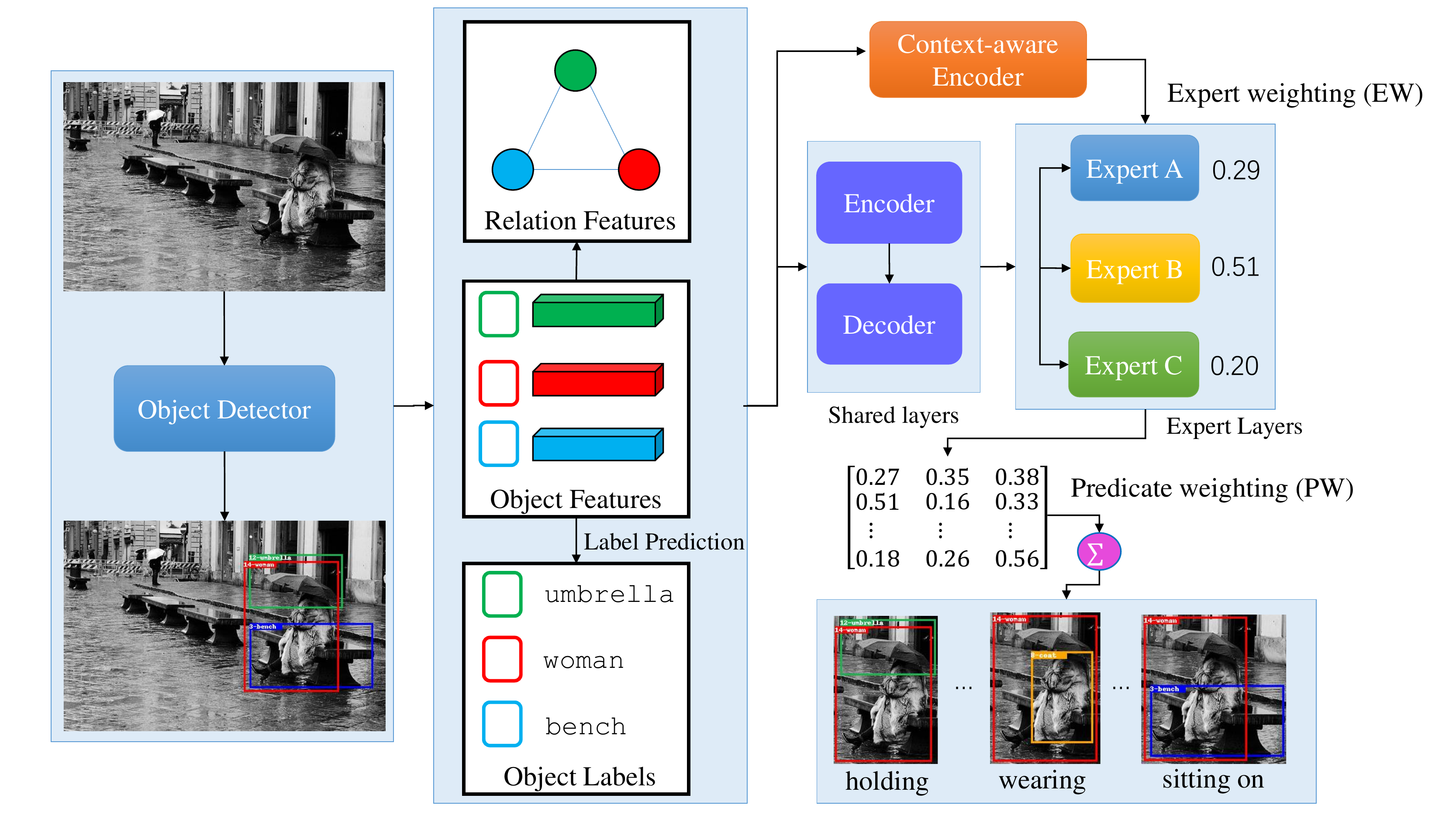}
	\end{minipage}}
	\caption{The framework of the proposed CAME for USGG is illustrated. We adopt the two-stage scene graph generation method. In the first stage, the object proposals, object labels, and relation features are estimated using the build-in SGG network $\operatorname{P}(B \mid I)$ and  $\operatorname{P}(O \mid B, I)$. In the second stage, for simplicity, we utilize the built-in context encoder as inputs for the context-aware encoder network. To construct the relation expert layers to expand the learning capabilities of original build-in scene graph generation methods, we utilize Mixture-of-Experts to build up $(P(R_i | B, O, I))$, including commonly shared layers of encoders/decoders network, and the final sets of expert layers, such as ``Classifier A''.}
	\label{fig:method}
\end{figure*}

In this section, we introduce the Context-Aware Mixture-of-Experts (CAME) for USGG. Our proposed framework is shown in Fig.~\ref{fig:method}. The SGG task can be decomposed into three tasks \cite{zellers2018neural,tang2020unbiased,chiou2021recovering}, including proposal generation, object classification, and relation prediction. Given an Image $\operatorname I$, the probability graph $\operatorname G$ obtained by SGG consists of a set of bounding boxes $\operatorname B$, object labels $\operatorname O$, and object pair relations $\operatorname R$,

\begin{equation}
	\operatorname{P}(G \mid I)=\operatorname{P}(B \mid I) \operatorname{P}(O \mid B, I) \operatorname{P}(R \mid B, O, I),
\end{equation}
where $\operatorname{P}(B \mid I)$ is the region proposals, mainly depending on the proposal generation network. In CAME, the region proposal network remains the same as the Faster R-CNN \cite{ren2015faster}. The $\operatorname{P}(O \mid B, I)$ is the object classification based on the region proposals. The ${P}(R \mid B, O, I)$ is the relation prediction based on images, object proposals, and object classifications. To achieve unbiased SGG, we mainly focus on the ${P}(R \mid B, O, I)$. There is an obvious contradiction between the limited learning capacity of a single model and the heavy long-tailed and diversified predicate distributions of the Visual Genome dataset. To solve this contradiction, 
we propose to utilize the mixture of experts to enlarge the learning capacity of one single model. The advantage is that we have multiple experts to handle the Visual Genome dataset, allowing each expert to learn the mutual complementary information under the guidance of context-aware loss, leading to a more diversified expert layer for long-tailed relation predicates classes. In detail, the SGG can be decomposed as:

\begin{equation}
	\operatorname{P}(G_i \mid I)=\operatorname{P}(B \mid I) \operatorname{P}(O \mid B, I) \operatorname{P}(R_i \mid B, O, I), i \in [1,n],
\end{equation}

where $n$ is the number of experts deployed in CAME. $\operatorname{P}(R_i \mid B, O, I)$ is the intermediate result obtained by i-th expert. $\operatorname{P}(G_i \mid I)$ is the prediction result obtained by i-th expert. The detailed construction of the expert layer in MOTIFS and VCTree is elaborated in supplementary material.

\begin{table*}[t]
	\scriptsize
	\renewcommand{\arraystretch}{1}
	\caption{SGG performs in percentage (\%) of various debiasing methods in the Visual Genome dataset. R@50/100 and mR@50/100 and their mean are reported over three tasks.}
	\label{tab:comparsion}
	\begin{center}
		\begin{tabular}{l|ccc|ccc|ccc}
			\hline
			& \multicolumn{3}{c}{PredCls} & \multicolumn{3}{c}{SGCls} & \multicolumn{3}{c}{SGGen}   \\ \hline
			Model		  & R@50/100 & mR@50/100 & mean & R@50/100 & mR@50/100 & mean & R@50/100 & mR@50/100  & mean	\\ \hline
			IMP \cite{xu2017scene,chen2019knowledge}       & 61.1 / 63.1 & 11.0 / 11.8 & 36.8 & 37.4 / 38.3 & 6.4 / 6.7   & 22.2 & 23.6 / 28.7 & 3.3 / 4.1   & 14.9 \\
			GPS-Net \cite{lin2020gps}    & 65.2 / 67.1 & 15.2 / 16.6 & 41.0 & 37.8 / 39.2 & 8.5 / 9.1   & 23.7 & 31.1 / 35.9 & 6.7 / 8.6   & 20.6 \\ 
			SG-CogTree \cite{yu2020cogtree} & 38.4 / 39.7 & 28.4 / 31.0 & 34.4 & 22.9 / 23.4 & 15.7 / 16.7 & 19.7 & 19.5 / 21.7 & 11.1 / 12.7 & 16.3 \\
			BGNN \cite{li2021bipartite}      & 59.2 / 61.3 & 30.4 / 32.9 & 46.0 & 37.4 / 38.5 & 14.3 / 16.5 & 26.7 & 31.0 / 35.8 & 10.7 / 12.6 & 22.5 \\ \hline
			%OSDI \cite{zhou2022debiased} & 28.5 & 33.6 & 35.5 & 16.6 & 20.4 & 21.4 & 7.7 & 10.3 & 12.1 \\ \hline
			%IMP+ \cite{xu2017scene,chen2019knowledge}            & - & 9.8 & 10.5 & - & 5.8 & 6.0 & - & 3.8 & 4.8  					\\
			%FREQ \cite{zellers2018neural,tang2019learning}	   & 8.3 & 13.0 & 16.0 & 5.1 & 7.2 & 8.5 & 4.5 & 6.1 & 7.1  			\\ 
			%KERN \cite{chen2019knowledge}
			%& - & 17.7 & 19.2 & - & 9.4 &  10.0 & - & 6.4 & 7.3 \\
			%GPS-Net \cite{lin2020gps}
			%& 17.4 & 21.3 & 22.8 & 10.0 & 11.8 & 12.6 & 6.9 & 8.7 & 9.8 \\
			%GB-Net \cite{zareian2020bridging}
			%& - & 22.1 & 24.0 & - & 12.7 & 13.4 & - & 7.1 & 8.5  \\ 	
			%Seq2Seq - RL \cite{lu2021context}
			%& 21.3 & 26.1 & 30.5 & 11.9 & 14.7 & 16.2 & 7.5 & 9.6 & 12.1	\\ 
			%OSDI \cite{zhou2022debiased} & 28.5 & 33.6 & 35.5 & 16.6 & 20.4 & 21.4 & 7.7 & 10.3 & 12.1 \\ \hline
			% 			Seq2Seq-RL \cite{lu2021context}       				   & 21.3 & 26.1 & 30.5 & 11.9 & 14.7 & 16.2 & 7.5 & 9.6 & 12.1 	\\ \hline
			%MOTIFS \cite{zellers2018neural}	   & 12.6 & 16.1 & 17.4 & 6.5 & 8.0 & 8.5 & 5.3 & 7.3 & 8.6  \\
			Motifs \cite{zellers2018neural} & 65.5 / 67.2 & 15.7 / 17.1 & 41.4 & 39.3 / 40.1 & 8.0 / 8.5 & 24.0  & 33.1 / 37.4 & 7.3 / 8.5 &   21.6 \\ \cline{2-10}
			%MOTIFS$\star$ 	   				   & 12.4 & 15.7 & 17.1 & 7.1 & 8.8 & 9.3 & 5.3 & 7.3 & 8.5  \\
			%MOTIFS-Focal$\star$				   & 12.2 & 16.2 & 18.2 & 7.0 & 8.9 & 9.7 & 4.5 & 6.4 & 7.7  \\ 
			%MOTIFS-LDAM$\star$				   & 12.6 & 15.9 & 17.2 & 7.2 & 8.9 & 9.5 & 5.2 & 7.0 & 8.2 	\\ 
			%MOTIFS-TDE (SUM)$\star$ \cite{tang2020unbiased}           & 16.3 & 22.9 & 26.9 & 10.2 & 13.7 & 15.6 & 6.7 & 9.1 & 10.9 		\\
			Motifs-TDE \cite{tang2020unbiased} 
			& 46.2 / 51.4 & 25.5 / 29.1 & 38.1 & 27.7 / 29.9 & 13.1 / 14.9 & 21.4 & 16.9 / 20.3 & 8.2 / 9.8 & 13.8 \\
			%MOTIFS-EBM \cite{suhail2021energy}     			   & 14.2 & 18.0 & 19.5 & 8.2 & 10.2 & 11.0 & 5.7 & 7.7 & 9.3 			\\ 
			%MOTIFS-Seg \cite{khandelwal2021segmentation}   	& 14.5 & 18.5 & 20.2 & 8.9 & 11.2 & 12.1 & 6.4 & 8.3 & 9.2  		\\ 
			% 			MOTIFS-ME-2				           & 17.8 & 25.4 & 31.1 & 9.4 & 13.6 & 16.2 & 6.6 & 9.4 & 11.8 & 	\\ 
			% 			MOTIFS-ME-3				           & 18.0 & 25.8 & 31.8 & 9.6 & 14.0 & 16.8 & 6.5 & 9.1 & 11.6 & 	\\ 
			% 			MOTIFS-ME-4		       	           & 17.8 & 26.1 & 31.7 & 9.6 & 13.9 & 16.5 & 6.4 & 9.1 & 11.5  & 	\\ 
			% 			MOTIFS-RAME-2                      & 19.1 & 27.2 & 32.3 & 9.9 & 14.7 & 17.4 & 6.4 & 9.2 & 11.7  & 	\\
			% 			MOTIFS-RAME-3                      & 17.9 & 25.7 & 31.5 & 9.4 & 14.2 & 17.1 &  &  &   & 	\\
			Motifs-CogTree \cite{yu2020cogtree}
			& 35.6 / 36.8 & 26.4 / 29.0 & 32.0 & 21.6 / 22.2 & 14.9 / 16.1 & 18.7 & 20.0 / 22.1 & 10.4 / 11.8 & 16.1 \\
			Motifs-RTPB	\cite{chen2022resistance}
			& 40.4 / 42.5 & 35.3 / 37.7 & 39.0 & 26.0 / 26.9 & 20.0 / 21.0 & 23.5 & 19.0 / 22.5 & 13.1 / 15.5 & 17.5 \\
			Motifs-PPDL	 \cite{li2022ppdl}
			& 47.2 / 47.6 & 32.2 / 33.3 & 40.1 & 28.4 / 29.3 & 17.5 / 18.2 & 23.4 & 21.2 / 23.9 & 11.4 / 13.5 & 17.5 \\ 
			Motifs-GCL	\cite{dong2022stacked}
			&  42.7 / 44.4 & {36.1} / {38.2} & 40.4 & 26.1 / 27.1 & \textbf{20.8} / \textbf{21.8} & 24.0 & 18.4 / 22.0 &  \textbf{16.8} / \textbf{19.3} & 19.1 \\ \cline{2-10}
			Motifs-CAME              & \textbf{55.3} / \textbf{57.4} & \textbf{37.9} / \textbf{40.1} & \textbf{47.7} & \textbf{34.6} / \textbf{35.5} & 19.3 / 21.3 & \textbf{27.7}  & \textbf{28.2} / \textbf{32.3} & 16.2 / 18.8 & \textbf{23.9} \\ \hline
			VCTree \cite{tang2019learning}	   & 65.9 / 67.5 & 17.2 / 18.5 & 42.3  & 45.3 / 46.2  & 10.6 / 11.3 & 28.4 & 31.9 / 36.2 & 7.1 / 8.3 & 20.9 			\\ \cline{2-10}
			%VCTree-Focal$\star$				   & 7.3  & 10.9 & 13.3 & 8.7 & 11.0& 11.9 & 4.3 & 6.3 & 7.6 		\\ 
			%VCTree-LDAM	$\star$				   & 7.5  & 10.7 & 12.8 & 7.9 & 9.5 & 10.1 & 4.0 & 5.3 & 6.1  			\\ 
			%VCTree-TDE(SUM)$\star$ \cite{tang2020unbiased}			& 19.5 & 26.2 & 29.8 & 10.5 & 15.0 & 17.4 & 6.9 & 9.6 & 11.5 	\\ 
			VCTree-TDE \cite{tang2020unbiased}
			& 47.2 / 51.6 & 25.4 / 28.7 & 38.2 & 25.4 / 27.9 & 12.2 / 14.0 & 19.9 & 19.4 / 23.2 & 9.3 / 11.1 & 15.8 \\ 
			% VCTree-EBM \cite{suhail2021energy}				 & 14.2 & 18.2 & 19.7 & 10.4 & 12.5 & 13.5 & 5.7 & 7.7 & 9.1   		\\         	% VCTree-Seg\cite{khandelwal2021segmentation}               &  15.0 & 19.2 & 21.1 &  9.3 & 11.6 & 12.3 & 6.3 & 8.1 & 9.0  	\\ 
			% 			VCTree-ME-2                        & 18.4 & 26.2 & 30.7 & 11.7 & 16.7 & 19.9 & 6.1 & 8.8 & 10.8  &		\\  
			% 			VCTree-ME-4                        & 18.1 & 25.8 & 30.6 & 12.0 & 17.2 & 20.6 & 5.3 & 7.7 & 10.0 &  		\\  	
			% 			VCTree-RAME-2						& 16.7 & 24.3 & 29.4 & 11.3 & 16.2 & 19.3 & 5.9 & 8.5 & 10.5 &	\\
			
			VCTree-CogTree \cite{yu2020cogtree}
			& 44.0 / 45.4 & 27.6 / 29.7 & 36.7 & 30.9 / 31.7 & 18.8 / 19.9 & 25.3 & 18.2 / 20.4 & 10.4 / 12.1 & 15.3 \\
			VCTree-RTPB	\cite{chen2022resistance} 
			& 41.2 / 43.3 & 33.4 / 35.6 & 38.4 & 28.7 / 30.0 &  \textbf{24.5} / \textbf{25.8} & 27.3 &  18.1 / 21.3 &  12.8 / 15.1 & 16.8		\\
			VCTree-PPDL	\cite{li2022ppdl}
			& 47.6 / 48.0 & 33.3 / 33.8 & 40.7 & 32.1 / 33.0 & 21.8 / 22.4 & 27.3 & 20.1 / 22.9 & 11.3 / 13.3 &  16.9 	\\
			VCTree-GCL	\cite{dong2022stacked}
			& 40.7 / 42.7 & {37.1} / {39.1} & 39.9 & 27.7 / 28.7 &  22.5 / 23.5 & 25.6 & 17.4 / 20.7 & 15.2 / 17.5 & 17.7 \\ \cline{2-10}
			VCTree-CAME
			& \textbf{55.5} / \textbf{57.4} & \textbf{39.4} / \textbf{41.5} & \textbf{48.5}  &  \textbf{38.6} /  \textbf{39.6} & 23.9 / 25.1 & \textbf{31.8} &  \textbf{27.0} /  \textbf{31.0} &  \textbf{16.0} / \textbf{18.9} & \textbf{23.2} \\ \hline 
			%				BGNN \cite{li2021bipartite}
			%				& - & 30.4 & 32.9 & - & 14.3 & 16.5 & - & 10.7 & 12.6 \\ 
			%				BGNN-CAME &  \textbf{26.7} & \textbf{33.6} & \textbf{36.4} & \textbf{15.7} & \textbf{19.7} & \textbf{21.5} & \textbf{9.6} & \textbf{13.8} & \textbf{16.7} \\ 
			%				BGNN \cite{li2021bipartite}
			%				& - & 30.4 & 32.9 & - & 14.3 & 16.5 & - & 10.7 & 12.6 \\ 
			%				BGNN-CAME &  \textbf{26.7} & \textbf{33.6} & \textbf{36.4} & \textbf{15.7} & \textbf{19.7} & \textbf{21.5} & \textbf{9.6} & \textbf{13.8} & \textbf{16.7} \\ 
			
			\hline
			Transformer \cite{li2021bipartite}
			& 65.5 / 67.2 & 17.7 / 19.1 & 42.4 & 40.3 / 41.1 & 10.4 / 11.0 & 25.7 & 32.7 / 37.1 & 8.2 / 9.6 & 21.9 \\ 
			Transformer-CAME &  55.1 / 57.2 & \textbf{37.4} /  \textbf{39.9} & \textbf{47.4} & 33.2 / 34.1 & \textbf{23.1} / \textbf{24.5} & \textbf{28.7} & 27.4 / 31.6 & \textbf{16.9} / \textbf{19.5} & \textbf{23.9} \\
			\hline

		\end{tabular}
	\end{center}
\end{table*}

\subsection{Mixture-of-Experts Layer}

The Visual Genome dataset contains the highly long-tailed distribution of predicates classes. When trained on the dataset, the model tends to be biased in predicting the head relationship and fails on predicting the body and tail relationship.

We propose to exploit the mixture of experts model with earlier layers as shared layers, which is denoted as $W_s$, and later a mixture of experts attached to the shared layers in parallel, which is denoted as $E_{i}$. Following the above-mentioned settings, the $\operatorname{P}(B \mid I)$ and
$\operatorname{P}(O \mid B, I)$ are considered as a shared layer $W_s$. Hence, the input image is processed by shared layer $W_s$ and the shared latent feature is obtained as $y_s$,

\begin{equation}
	y_{s}=Ws(x).
\end{equation}

Specifically, $y_s$ of MOTIFS contains rich context information of the scene.
The $\operatorname{P}(R_i \mid B, O, I)$ is considered as the expert layer.
Then, the shared latent feature $y_s$ is processed by each expert $E_i$ of the expert layer individually as

\begin{equation}
	y_{i}= E_i (y_s), i \in [1,n],
\end{equation}

\begin{equation}
	y_{moe}=\frac{1}{n} \sum_{i=1}^{n} y_i,
\end{equation}
where $n$ is the number of experts deployed in the network architecture. Empirically, there are sets of experts used in network architecture. The output of the model is the average of all experts as $y_{moe}$. The shared layers and expert layers are jointly trained in the training stage. At the inference stage, the results from all expert layers are averaged in the logits form for the final ensemble.

\subsection{Context-aware Mixture-of-Experts}
\label{sec:CAME}
\textbf{Expert weighting:}
The traditional Mixture-of-Experts model lacks the context information of the input images and gives each expert essentially the same weight. Besides, some adopt a gating strategy to expedite the inference process, which abandons some experts in a gating mechanism, resulting in a decline in accuracy. To overcome the aforementioned shortcomings, as seen in Fig.~\ref{fig:ca_ew_pw}, we propose the unique context-aware expert weighting scheme that dynamically assigns the relevance of each expert based on scene context without compromising performance. In the meanwhile, it boosts the model's capacity for unbiased SGG and its diversity.

%	\begin{figure*}[thpb]
%	\begin{center}
%	\begin{tabular}{ccc}
%		\includegraphics[width=5cm]{Figs/head_improve.png} &
%		\includegraphics[width=5cm]{Figs/body_improve.png} & 
%		\includegraphics[width=5cm]{Figs/tail_improve.png} 
%	\end{tabular}
%	\caption{The class-wise re-weighting parameters of head, body, and tail classes.}
%	\label{fig:visualzation_gamma} 
%	\end{center}
%	\end{figure*}

There are primarily two important steps in the SGG approach. First is object context processing, and then is edge context processing. In this paper, object context processing is utilized to build the context encoder. Instead of building context encoders in our own way, we borrow the original baseline models that have a built-in context encoder as a shared context encoder. For example, in the MOTIFS \cite{zellers2018neural}, the context encoder is a bidirectional LSTM while, in the VCTree \cite{tang2019learning}, the context encoder is constructed with a bidirectional TreeLSTM. In the MOTIFS and VCTree, the context information includes the feature of object proposals, the position of object proposals, feature of object labels. We presume that the object context information provides a strong indication of its intrinsic predicates.

%\begin{figure*}[thpb]
%	\begin{center}
%		\begin{tabular}{c}
%			\includegraphics[width=18cm,height=6cm]{Figs/per_cls_came3mot3.png} %&
%			%				\includegraphics[width=4cm]{Figs/body_improve.png} & 
%			%				\includegraphics[width=4cm]{Figs/tail_improve.png} 
%		\end{tabular}
%		\caption{The per-class Recall@100 of PredCls difference between the MOTIFS-CAME and MOTIFIS over the body and tail classes.}
%		\label{fig:visualization_predcls} 
%	\end{center}
%\end{figure*}

As shown in Fig.~\ref{fig:ca_ew_pw}, these object information will be passed into bidirectional LSTMs first, and then the relational characteristics between these objects are obtained with such abundant input features. The relational feature is aggregated and passed through an MLP layer $W_{context}$ with softmax attached to the output. Consequently, the normalized significance of each expert is obtained as $\beta_i$,

\begin{equation}
	\beta = W_{context} (B, O, I), \beta \in \mathbb{R}^{n}
\end{equation}

\begin{equation}
	\beta_{i}=\frac{\exp{  (W_{context}(y_i)) }}{\sum_{i=1}^{n} \exp {  (W_{context}(y_i)) }}
\end{equation}

%{\textbf{Re-weighting Loss Factor:}} As shown in Fig.~\ref{fig:data_distribution}, the data distribution of head, body and tail classes are heavily biased. To remedy this data distribution, re-weighting loss is applied to different predicates as:
%\begin{equation}
%	L_{re}=\sum_{i=1}^{n} \gamma_{i} L_{cls} ({x,f(x, \theta_{i}) }) 
%\end{equation}
%
%where $\gamma_{i}$ is the re-weighted importance of each predicate. We utilize the effective numbers as weighting factor \cite{cui2019class}:
%
%\begin{equation}
%	\gamma_{i}=\frac{1-\beta}{1-\beta^{n_k}}
%\end{equation}
%
%where $n_k$ is number of samples for each class, and we use $\beta$ as 0.9999 in experiments. With this reweighting loss, the long-tailed problem is mitigated. $\gamma_{i}$ regulates the importance of each predicate class, i.e., if the sample belongs to head classes, $\gamma_{i}$ further reduces the weights of sample, because the dominant predicate classes may overfit the classifier to biased classifier. Meanwhile, if the sample belongs to tail classes, $\gamma_{i}$ will be larger than head classes, and loss is not expanded too much since the data of tail classes are relatively limited. 

\textbf{Predicate weighting:}
In addition to adding specific weights to each expert, we propose a predicate weighting strategy to find the relevance of each predicate to each expert. The predicate prediction of each expert contains the probability for a series of predicate classes, which could be further exploited to increase the diversity of model. Therefore, we utilize a built-in edge processing module as a context encoder, to further increase the learning capacity as well as variety of model. Specifically, to establish relationship between predicate prediction to context information, the relation distribution after edge context processing is utilized as $y_e$. This information will be processed by relation encoder $W_{Rel_{context}}$ to obtain relational weighting between each expert.  $r_k$ is of dimension $m\times n$. The normalized significance of each relation of i-th expert is obtained as $r_{{k}_{i}}$ with a size of $m \times 1$, 

%and the original probabilities are summed by accumulating the predictions of each expert, which further increases the diversity of expert layers
\begin{equation}
	r_k = \gamma * W_{Rel_{context}} (y_e), 
\end{equation}

\begin{equation}
	r_{{k}_{i}}=\frac{\exp{  (W_{Rel_{context}}(y_e)) }}{\sum_{i=1}^{n} \exp {  (W_{Rel_{context}}(y_e)) }}
\end{equation}

where $\gamma$ is a hyperparameter between [0,10], and m is the number of relations in the prediction result. 

\textbf{Context-aware Loss Function:} Based on a set of context information, different experts are sensitive to different predicate data distributions. For instance, expert A could be more sensitive to predicates from head classes, while expert C could be more sensitive to predicates from tail classes. Therefore, the novel context-aware loss $\mathcal{L}_{\text{ca }}$ is designed to increase the diversity of each expert, therefore to better handle such scenarios. These $n$ experts are jointly trained on context-aware classification loss, where the base classification loss $\mathcal{L}_{\text{Cls}}$ could be LDAM \cite{cao2019learning}, class balanced loss \cite{cui2019class}, CE, and focal \cite{lin2017focal} loss.

\begin{equation}
	\mathcal{L}_{\text{ca }}(x, y)=\sum_{i=1}^{n}\left(\beta_i \mathcal{L}_{\text{Cls }}\left(x, y_i ; \theta_{i}\right) )\right)
\end{equation}

Then, the ensemble output of the CAME network is formulated as $y_{came}$,
\begin{equation}
	y_{came}=\frac{1}{n} \sum_{i=1}^{N} r_{{m}_{i}} y_{i}
\end{equation}

where $y_{i}$ is $m\times 1$ vector contains the predicate prediction of i-th expert.
% \begin{equation}
% \lambda \cdot\mathcal{L}_{\text {D-Diversify }}\left(x, y ; \theta_{i}\right
% \end{equation}

\section{Experiments and Results}

\textbf{Visual Genome:}
For the SGG task, the Visual Genome (VG) dataset is used to train and evaluate our proposed models. VG dataset \cite{krishna2017visual} is a large-scale visual dataset, and consists of the components of region descriptions, objects, attributes, relationships, region graphs, scene graphs, and question answer pairs. VG has been widely used for scene graph generation and its applications. VG dataset contains over 108K images and has an average of 21 objects, 18 attributes, and 18 pairwise relationships between objects in each image. In this paper, we use the most widely used settings \cite{tang2019learning,tang2020unbiased,chiou2021recovering} that the most frequent 50 predicates and 150 objects are preserved for training and testing. The setting contains 62,723 images as the training set, 5,000 images selected from the training set as the validation set, and 26,446 images as the test set.

\textbf{Scene Graph Generation Models:}
The CAME is generic and does not require any additional information on the backbone of scene graph generation models. This feature allows the free integration and enhancement of any existing scene graph generation architectures. Since the CAME is a general framework that could be integrated into various SGG models, we mainly select two representative models, i.e. Neural MOTIFS and VCTree, which have been widely used to compare and validate the performance of various methods \cite{khandelwal2021segmentation, guo2021general,suhail2021energy,tang2020unbiased} on the Visual Genome dataset.

\subsection{Implementation Details}

\textbf{Object Detection:}
For object detection, the two-stage detector Faster R-CNN \cite{ren2015faster} is used, where the backbone model is ResNetXt101-FPN \cite{Xie2016}. The weight of pretrained detection model for the Visual Genome dataset is obtained from the open-source scene graph generation project \cite{tang2020sggcode}. The mAP of the backbone for object detection on the Visual Genome dataset is 0.28.

\textbf{CAME:}
For the training of CAME on unbiased SGG task, the batch size during training is 12 with two NVIDIA RTX 2080 Ti. The learning rate is 0.01 for the Motifs and VCTree baseline, and 0.001 for the Transformer baseline, with a warmup factor of 0.1. The weight decay is 0.0001 whereas the momentum is 0.9. The grad norm clip is 5.0. The number of experts is ranging from 2 to 4.

\subsection{Evaluation}

\textbf{Performance Metric:}
Same as \cite{tang2020unbiased}, we use the Top-K recall to measure the system performance. The Top-K recall calculates the fractions of the ground truth instances hit in the top-K predictions in each class, where K is 20, 50, or 100. Therefore, R@50, and R@100 are obtained. However, as the recall could have been dominated by most head classes like \textbf{on} predicted by the biased model, we follow \cite{tang2020unbiased,chiou2021recovering,khandelwal2021segmentation,suhail2021energy} to average the recalls of each class over the whole dataset, and utilize the less-biased performance metric mean Recall, including mR@50 and mR@100. Nevertheless, we observe there is a trade-off between Recall and Mean Recall, i.e.,  increasing the mean Recall leads to a tremendous dropping of Recall metrics in recent works like GCL, PPDL, and RTPB \cite{dong2022stacked,li2022ppdl,chen2022resistance}. To comprehensively measure the USGG, we take the mean of R@50/100, and mR@50/100, as an essential balance metric in USGG. Moreover, to ensure a balanced SGG, we propose to use variance of recalls over mean of recalls, denoted as $\mathbf{var/m}$, to show the balance between various classes.

%R@20, 
%mR@20, 
Scene graph generation mainly focuses on the localization of objects and classification of $	\left \langle subject, predicate, object \right \rangle $ triplet. We follow three conventional metrics to evaluate our SGG model: (1) \textbf{Scene Graph Generation (SGGen)}: This task is to detect a set of predefined objects and predict the relation between each object pair.  (2) \textbf{Scene Graph Classification (SGCls)}: The goal of this task is to predict the class labels for the set of objects and the relationship among each object pair. (3) \textbf{Predicate Classification (PredCls)}: This task is to predict the relationship predicate among all the object pairs. It is evaluated by the accuracy of predicate classification.

The aforementioned three tasks are used to assess CAME. Besides, following the setting of long-tailed distribution \cite{liu2019large, chiou2021recovering} and according to the number of each predicate in the dataset, we divide the distribution of predicate classes into three categories, including head classes (top-15 predicates), body classes (middle-20 predicates), and tail classes (last-15 predicates). To validate the effectiveness of the proposed method on the infrequent predicates, the mean Recall on body and tail classes are presented.

\begin{figure}[ht!]
	\begin{center}
		%\resizebox{\columnwidth}{!}{
		\begin{tabular}{c}
			\includegraphics[width=8cm]{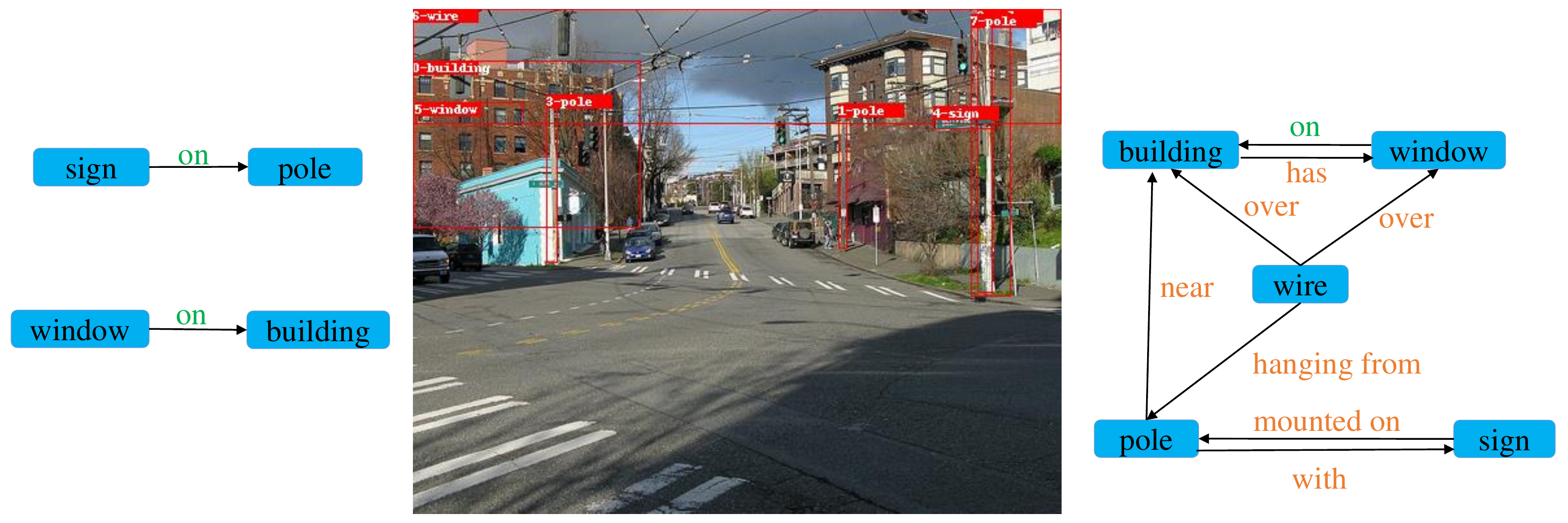}    \\
			a)												\\
			\includegraphics[width=8cm]{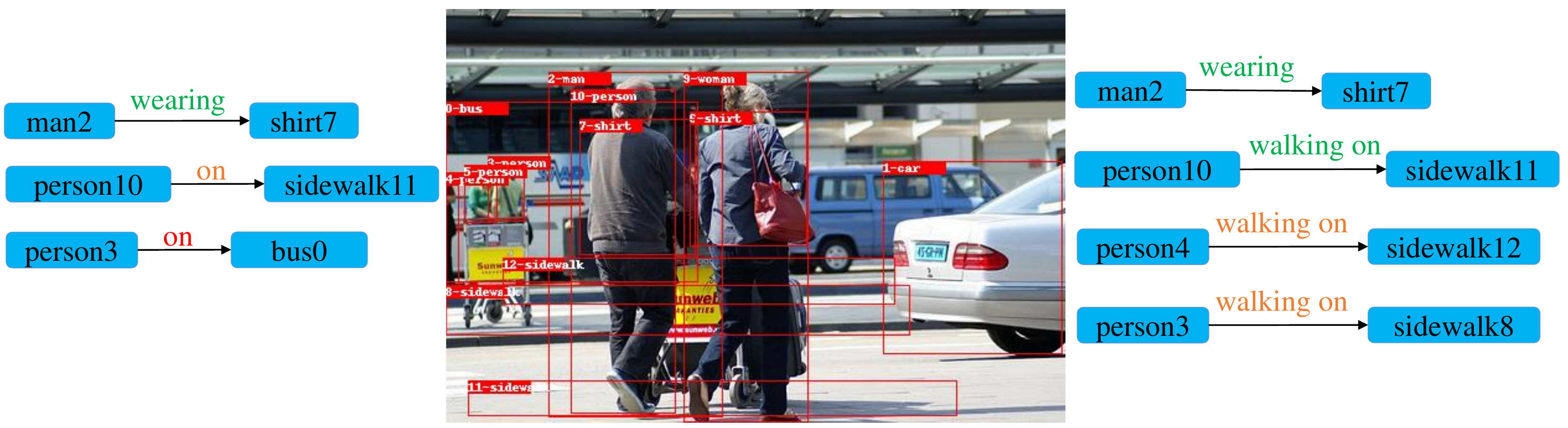}   	\\
			b)												\\
			\includegraphics[width=8cm]{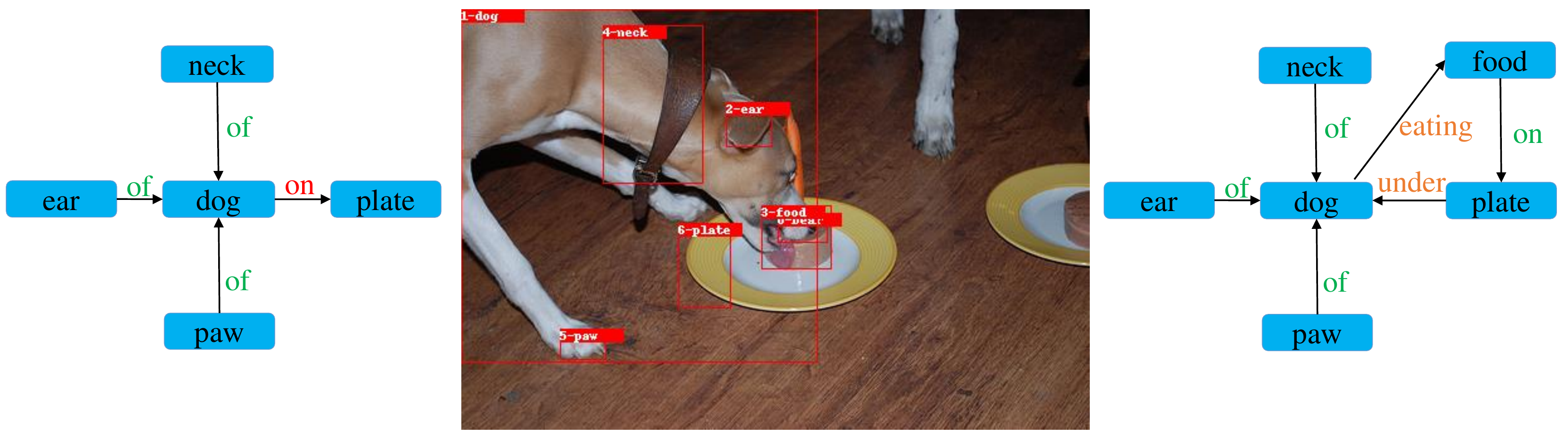}  	\\ 
			c)												\\
		\end{tabular}
		%}
		\caption{Qualitative examples of SGG by MOTIFS and MOTIFS-CAME in PredCls setting. The left prediction is estimated by MOTIFS, and the right prediction is estimated by MOTIFS-CAME. We show the prediction with the highest confidence, where the \textcolor{green}{correct} (the same with GT),  \textcolor{red}{incorrect} (not match the GT and not appropriate), or  \textcolor{brown}{acceptable} (not match the GT but actually reasonable.} 
		\label{fig:visualization_sgg} 
	\end{center}
\end{figure}

%\begin{figure*}[thpb]
%	\begin{center}
%		\begin{tabular}{cc}
%			\includegraphics[width=7cm]{Figs/Fig_5_1.pdf} &
%			\includegraphics[width=7cm]{Figs/Fig_5_2.pdf}  \\
%			a) &
%			b)  \\
%			\includegraphics[height=5cm]{Figs/Fig_5_3.pdf} &
%			\includegraphics[height=5cm]{Figs/Fig_5_4.pdf} \\
%			c) &
%			d) \\
%		\end{tabular}
%		\caption{Qualitative examples of SGG by MOTIFS and MOTIFS-CAME in PredCls setting. The object with the blue bounding box is the subject and with green bounding box is the object. The top prediction is estimated by MOTIFS, and the bottom prediction is estimated by MOTIFS-CAME. We show the top-5 classes of highest confidences for predicate prediction, where the \textcolor{green}{correct} (the same with GT),  \textcolor{red}{incorrect} (not match the GT and not appropriate), or  \textcolor{brown}{acceptable} (not match the GT but actually reasonable.} 
%		\label{fig:visualization_prob} 
%	\end{center}
%\end{figure*}

\textbf{Comparisons with SOTA Models:}
%GPS-Net \cite{lin2020gps} and GB-Net \cite{zareian2020bridging}
%energy-based loss \cite{suhail2021energy}, and Segmentation grounded \cite{khandelwal2021segmentation}
To demonstrate the efficacy of CAME, we first compare it to previous biased SGG approaches such as Motifs \cite{zellers2018neural}, VCTree \cite{tang2019learning}, and Transformer. In particular, we primarily compare CAME with unbiased SGG approaches, containing the TDE \cite{tang2020unbiased}, CogTree \cite{yu2020cogtree}, RTPB \cite{chen2022resistance}, GCL \cite{dong2022stacked} and PPDL \cite{li2022ppdl} on USGG approaches on the MOTIFS and VCTree baselines.

The experiment results of unbiased scene graph generation are shown in Table~\ref{tab:comparsion}. In comparison with the baseline methods, the proposed CAME method reaches the best performance on the mean Recall, with a relative improvement of \textbf{235.9\%} over the baseline MOTIFS and \textbf{225.8\%} over the baseline VCTree, indicating that the CAME achieves a notable performance improvement over the biased SGG methods. Besides, the mean of baseline models is superior to the majority of existing approaches, indicating that the latter are not properly debiasing, but rather transfer their weights from head to tail classes.

Meanwhile, on the PredCls task, in comparison with SOTA debiasing methods, the proposed CAME surpasses the existing approaches, with a relative improvement of \textbf{15.7} and \textbf{7.3}  over the MOTIFS-CogTree and MOTIFS-GCL on the MOTIFS structure. Besides, on the VCTree baseline, CAME surpasses the VCTree-CogTree and VCTree-GCL, with a relative improvement of \textbf{11.8} and \textbf{8.6}, respectively. 

In conclusion, our methods can improve the USGG ability on the mean recalls and obtain well-preserved recalls.  In contrast, recent studies exploit the limits on the mean recalls,  leading to a biased predictor on the tail predicates, while neglecting the overall USGG performance on the recalls. Hence, there is a boost in the mean recall metric mR@50/100 while an enormous degradation on the R@50/100. We believe these two metrics are of equal importance, hence, the mean of these two metrics tips the balance of USGG to some degree, avoiding the heavily biased predictors for the head classes. We find that recent studies are heavily biased to the tail classes on the conditional of sacrificing performance on the head classes, leading to an inferior performance on the mean of two metrics. Instead, in terms of both recall and mean recall, we have obtained satisfactory results without much shifting the model's cognition ability to the tail classes. For instance, we have obtained \textbf{47.7} and \textbf{48.5} on the PredCls task on the Motifs and VCTree baselines, while the second best among these unbiased SGG approaches are 40.4 and 40.7, which shows an 18.1\% and 19.2\%  improvement over recent methods.

\subsection{Ablation Studies}
\textbf{Comparison with other imbalance strategies:}
To show the difference and superiority of CAME, there are various losses used for long-tailed problem are compared, as shown in Table~\ref{tab:predcls}. The performance of the Motifs baseline model, LDAM loss, focal loss, TDE, and CAME are presented under the PredCls settings. The results show that the CAME reaches a notable performance boost compared with other debiasing methods. For example, in comparison with the TDE methods, CAME further improves mean recalls on the body classes about \textbf{12.0\%}. Notably, CAME is about \textbf{6} times compared with TDE on the prediction of tail classes. Meanwhile, the variance over mean of tail classes is relatively small \textbf{12.6\%}, which is around 1/4 of TDE. In total, these results show the superior of CAME on the unbiased SGG. The smaller variance shows that the CAME performs more equally with less bias over different classes.

%Table~\ref{tab:sggcls} illustrates the performance of CAME and other debiasing methods in SGCls settings. The CAME outperforms the TDE over \textbf{5.5\%}. Notably, the mean recall of tail classes is 11.1, which is around \textbf{9} times compared with TDE, and \textbf{23} times compared with baseline methods. Besides, the variance of tail classes is smaller than TDE. These results indicate the superior of CAME on the long-tail predicates prediction. The smaller variance shows that the CAME performs more equally with less bias over different classes. In summary, CAME has a higher unbiased SGG ability. 

\begin{table}[]
	\centering
	\footnotesize
	\caption{The SGG performance in PredCls settings. We show the mR@100, and performance on body and tail classes and its variance over mean recalls in percentage  with MOTIFS baseline.}
	\label{tab:predcls}
	\begin{tabular}{l|cc|cccc}
		\hline
		&  mR@100  & var/m & body & var/m  & tail & var/m \\ \hline
		Baseline & 17.7     & 36.8 & 7.8 & 13.0 & 2.0 & 25.5 \\
		LDAM     & 17.2     & 36.8 & 7.7 & 12.6 & 2.0 & 25.1 \\ 
		Focal    & 18.2     & 33.8 & 12.3 & 27.4 & 3.3 & 38.0  \\ 
		TDE      & 27.3     & 33.8 & 30.9 & 28.7 & 3.7 & 51.5 \\ \hline
		CAME & \textbf{40.1}& \textbf{14.0} & \textbf{38.8} & 15.8 & \textbf{39.9} & \textbf{7.6} \\ \hline \hline
	\end{tabular}
\end{table}

\textbf{The number of mixture of experts:}
We have evaluated the proposed method on the MOTIFS baseline for comparison with different settings. The findings of the ablation study are displayed in Table~\ref{tab:ablation_num_experts}. The ME and EW module are used for training. To demonstrate the impact of varying numbers of experts, the ME module is included with the number of experts ranging from 2 to 4 in order to provide the influence of number of experts. We note that the model with three experts performs the best. Therefore, the model described in this study includes three experts by default.

\begin{table}[ht]
	\footnotesize
	\centering
	\caption{Ablation for the PredCls task with ME and EW is reported. MOTIFS is base model. The ME is ranging from 2 to 4.}
	\label{tab:ablation_num_experts}
	\begin{tabular}{cc|ccc}
		\hline
		%\multicolumn{2}{c}{Configs}  & \multicolumn{3}{c}{PredCls}        \\ \hline
		ME & EW      & mR@50/100 & R@50/100 & Mean \\ \hline
		2 & \Checkmark  & 34.3 / 36.5 & 56.7 / 58.7 & 46.6   \\
		3 & \Checkmark  & 37.9 / 40.1 & 55.3 / 57.4 & 47.7  \\
		4 & \Checkmark  & 36.7 / 39.1 & 55.8 / 57.8 & 47.4 \\ \hline
		\hline \hline
	\end{tabular}
\end{table}

\textbf{The influence of different modules:}
As shown in Table \ref{tab:ablation_modules_mot} and \ref{tab:ablation_modules_vc}, the ME module first establishes a solid basis for unbiased SGG. Consequently, the EW module significantly improves the efficacy of USGG. For instance, it demonstrates a substantial improvement in the R@50, and R@100 on the PredCls task, while preserving the  mR@50, and mR@100. We have obtained \textbf{47.7} and \textbf{48.1} Mean on the PredCls for the Motifs and VCTree baselines, with the EA module attaining a relative improvement of \textbf{2.1} and \textbf{2.0} above ME. Moreover, the PW module continues to improve the performance of USGG to \textbf{49.0} with \textbf{58.3} and \textbf{60.2} on R@50/100 with VCTree baseline.

%ME module with the context information on the mR@50, and mR@100, but also shows a tremendous improvement on the R@50, and R@100 on PredCls task. In terms of these metrics, we have obtained satisfactory improvement without much shifting the model's cognition ability to the tail classes. For instance, we have obtained \textbf{57.1} and \textbf{57.3} of R@100 on the PredCls on the Motifs and VCTree baselines, CA module achieving a relative improvement of \textbf{2.1} over ME on the MOTIFS. The experiments validate the effectiveness of CAME on the unbiased SGG task. Moreover, the PCA module continue to improves the performance of module to \textbf{56.8} and \textbf{58.9}.

%Specifically, our methods can improve the USGG ability on the mean recalls and obtain well-preserved recalls. We believe these two metrics are of equal importance, hence, the mean of these two metrics tips the balance of USGG to some degree, avoiding the heavily biased predictors for the head classes. We measure the performance on the M@50, and M@100. 

\begin{table}[ht]
	\footnotesize
	\centering
	\caption{Ablation for PredCls task with designed modules is reported. Motifs is the base architecture for all methods.}
	\label{tab:ablation_modules_mot}
	\begin{tabular}{l|ccc}
		\hline
		%\multicolumn{2}{c}{Configs}  & \multicolumn{3}{c}{PredCls}     \\ \hline
		& mR@50/100 & R@50/100 & Mean \\ \hline
		Baseline  &    15.7 / 17.1 & 65.5 / 67.2 & 41.4  \\ \hline
		ME        &    37.7 / 40.3 & 51.2 / 53.3 & 45.6    \\
		ME+EW     &    37.9 / 40.1 & 55.3 / 57.4 & 47.7  \\
		ME+EW+PW  &    36.9 / 39.3 & 56.8 / 58.9 & 48.0 \\
		\hline \hline
	\end{tabular}
\end{table}

\begin{table}[ht]
	\footnotesize
	\centering
	\caption{Ablation for PredCls task with designed modules is reported. VCTree is the base architecture for all methods.}
	\label{tab:ablation_modules_vc}
	\begin{tabular}{l|ccc}
		\hline
		%\multicolumn{1}{c}{Configs}  & \multicolumn{3}{c}{PredCls}     \\ \hline
		& mR@50/100    & R@50/100    & Mean  \\ \hline
		Baseline    & 17.1 / 18.4  & 65.9 / 67.5 & 42.2  \\ \hline
		ME       & 38.9 / 41.2  & 51.1 / 53.0 & 46.1  \\
		ME+EW    &  38.5 / 41.1 & 55.4 / 57.4 & 48.1  \\
		ME+EW+PW &  37.4 / 40.0 & 58.3 / 60.2 & 49.0 \\ 
		\hline \hline
	\end{tabular}
\end{table}

\textbf{Hyperparameters selection:}
To demonstrate the efficiency of the predicate weighting module, a number of hyperparameters were utilized. As demonstrated in Table \ref{tab:ablation_hyper}, the optimal performance of the model is achieved when the hyperparameter $\alpha$ is 0.25, displaying \textbf{58.3} and \textbf{60.2} on R@50/100 while maintaining \textbf{37.4} and \textbf{40.0} on mR@50/100.

\begin{table}[ht]
	\footnotesize
	\centering
	\caption{Ablation with a series of $\alpha$ for PredCls task is reported. VCTree is the base architecture.}
	\label{tab:ablation_hyper}
	\begin{tabular}{c|ccc}
		\hline
		%\multicolumn{1}{c}{Configs}  & \multicolumn{3}{c}{PredCls}     \\ \hline
		$\alpha$ & mR@50/100 & R@50/100 & Mean \\ \hline
		0.25 &  37.4 / 40.0 & 58.3 / 60.2 & 49.0 \\ 
		0.5  &  38.9 / 41.3 & 56.6 / 58.5 & 48.8 \\
		1    &  38.8 / 41.6 & 55.7 / 57.7 & 48.5 \\
		5    &  38.8 / 40.8 & 55.8 / 57.7 & 48.3 \\
		\hline \hline
	\end{tabular}
\end{table}

%We also compare with common imbalance strategies, including conventional strategies including Focal, equalization loss (EQL), Resample, and Reweight, as shown in Table ~\ref{tab:imb_comparsion}. We observe that these losses do not perform well on the mean recalls, e.g., Focal performs slightly worse than baseline, while EQL, Resample, and Reweight are only slightly better than baseline. In detail, Resample and Reweight have achieved 3.33 and 3.65 relative improvements in mean metric. CAME has achieved 11.65 relative improvements in mean metric, which is about 2 times than baseline methods.

\textbf{Qualitative Analysis}
Fig.~\ref{fig:visualization_sgg} shows SGG by MOTIFS and MOTIFS-CAME to demonstrate the efficiency of the CAME. a) shows that $\textbf{sign-on-pole}$, $\textbf{window-on-building}$ predicted by MOTIFS match the ground truth but are less descriptive. In contrast, $\textbf{sign-mounted on-pole}$, $\textbf{wire-hanging from-pole}$, $\textbf{wire-over-building}$ and $\textbf{pole-near-building}$ are more expressive and valuable, though the ground truth does not contain such informative information. b) $\textbf{dog-on-plate}$ is incorrect, while $\textbf{plate-under-dog}$ and $\textbf{dog-eating-food}$ are acceptable. However, due to the missing labels or unlabeled predicates, such important information are missing. c) shows that the \textbf{man-wearing-shirt} matches the ground truth, $\textbf{person-on-sidewalk}$ is acceptable but does not match the ground truth, and $\textbf{person-on-bus}$ is wrong since the person is on the sidewalk. In contrast, the $\textbf{person-walking on-sidewalk}$ is more descriptive and matches the ground truth, where the \textbf{walking on} belongs to tail classes while \textbf{on} belongs to head classes. These samples indicate that the more valuable and expressive predicates could be estimated with the proposed CAME, although there are plenty of missing predicates in dataset.

\section{Conclusion}

In this paper, we propose a simple and effective Context-Aware Mixture-of-Experts (CAME) framework for unbiased SGG, which could be applied to various SGG benchmarks. The mixture of experts is utilized for unbiased SGG, with each expert learning the mutually complementary knowledge of predicate classes under the guidance of context-aware loss. Hence, the expert layer is able to obtain more comprehensive knowledge of predicate classes. Hence, the proposed CAME improves the model's unbiased SGG ability with complementary information of predicates learned by the expert network. Meanwhile, the SGG generation ability is well-preserved when improving the USGG. Moreover, the context-aware loss further enhances the relation between context information and each expert, leading to a more generalized and diversified model.

%This could be accomplished in two steps. First, expert layers are constructed based on a shared common encoder/decoder network and particular expert layers, such as relation classifiers, which significantly enhances the learning capacity of the model for long-tailed predicate distributions. Since each expert shares the same importance for various predicate classes, which is not plausible in actual cases, we propose a context-aware mixture of experts to dynamically assign the importance of each expert with the input context. Besides,  a novel context-aware loss is introduced to increase the diversity of the proposed expert network. As a result,
%\newpage
%%%%%%%%% REFERENCES
{\small
	\bibliographystyle{ieee_fullname}
	\bibliography{Ref}

\begin{thebibliography}{10}\itemsep=-1pt

\bibitem{armeni20193d}
Iro Armeni, Zhi-Yang He, JunYoung Gwak, Amir~R Zamir, Martin Fischer, Jitendra
  Malik, and Silvio Savarese.
\newblock 3d scene graph: A structure for unified semantics, 3d space, and
  camera.
\newblock In {\em Proceedings of the IEEE/CVF International Conference on
  Computer Vision}, pages 5664--5673, 2019.

\bibitem{cao2019learning}
Kaidi Cao, Colin Wei, Adrien Gaidon, Nikos Arechiga, and Tengyu Ma.
\newblock Learning imbalanced datasets with label-distribution-aware margin
  loss.
\newblock {\em Advances in neural information processing systems}, 32, 2019.

\bibitem{chen2022resistance}
Chao Chen, Yibing Zhan, Baosheng Yu, Liu Liu, Yong Luo, and Bo Du.
\newblock Resistance training using prior bias: toward unbiased scene graph
  generation.
\newblock {\em arXiv preprint arXiv:2201.06794}, 2022.

\bibitem{chen2019knowledge}
Tianshui Chen, Weihao Yu, Riquan Chen, and Liang Lin.
\newblock Knowledge-embedded routing network for scene graph generation.
\newblock In {\em Proceedings of the IEEE Conference on Computer Vision and
  Pattern Recognition}, pages 6163--6171, 2019.

\bibitem{chiou2021recovering}
Meng-Jiun Chiou, Henghui Ding, Hanshu Yan, Changhu Wang, Roger Zimmermann, and
  Jiashi Feng.
\newblock Recovering the unbiased scene graphs from the biased ones.
\newblock In {\em Proceedings of the 29th ACM International Conference on
  Multimedia}, pages 1581--1590, 2021.

\bibitem{cong2022reltr}
Yuren Cong, Michael~Ying Yang, and Bodo Rosenhahn.
\newblock Reltr: Relation transformer for scene graph generation.
\newblock {\em arXiv preprint arXiv:2201.11460}, 2022.

\bibitem{cui2019class}
Yin Cui, Menglin Jia, Tsung-Yi Lin, Yang Song, and Serge Belongie.
\newblock Class-balanced loss based on effective number of samples.
\newblock In {\em Proceedings of the IEEE/CVF conference on computer vision and
  pattern recognition}, pages 9268--9277, 2019.

\bibitem{dai2021generalizable}
Yongxing Dai, Xiaotong Li, Jun Liu, Zekun Tong, and Ling-Yu Duan.
\newblock Generalizable person re-identification with relevance-aware mixture
  of experts.
\newblock In {\em Proceedings of the IEEE/CVF Conference on Computer Vision and
  Pattern Recognition}, pages 16145--16154, 2021.

\bibitem{dhamo2020semantic}
Helisa Dhamo, Azade Farshad, Iro Laina, Nassir Navab, Gregory~D Hager, Federico
  Tombari, and Christian Rupprecht.
\newblock Semantic image manipulation using scene graphs.
\newblock In {\em Proceedings of the IEEE/CVF Conference on Computer Vision and
  Pattern Recognition}, pages 5213--5222, 2020.

\bibitem{dong2022stacked}
Xingning Dong, Tian Gan, Xuemeng Song, Jianlong Wu, Yuan Cheng, and Liqiang
  Nie.
\newblock Stacked hybrid-attention and group collaborative learning for
  unbiased scene graph generation.
\newblock In {\em Proceedings of the IEEE/CVF Conference on Computer Vision and
  Pattern Recognition}, pages 19427--19436, 2022.

\bibitem{gu2019unpaired}
Jiuxiang Gu, Shafiq Joty, Jianfei Cai, Handong Zhao, Xu Yang, and Gang Wang.
\newblock Unpaired image captioning via scene graph alignments.
\newblock In {\em Proceedings of the IEEE/CVF International Conference on
  Computer Vision}, pages 10323--10332, 2019.

\bibitem{gu2019scene}
Jiuxiang Gu, Handong Zhao, Zhe Lin, Sheng Li, Jianfei Cai, and Mingyang Ling.
\newblock Scene graph generation with external knowledge and image
  reconstruction.
\newblock In {\em Proceedings of the IEEE Conference on Computer Vision and
  Pattern Recognition}, pages 1969--1978, 2019.

\bibitem{guo2021general}
Yuyu Guo, Lianli Gao, Xuanhan Wang, Yuxuan Hu, Xing Xu, Xu Lu, Heng~Tao Shen,
  and Jingkuan Song.
\newblock From general to specific: Informative scene graph generation via
  balance adjustment.
\newblock In {\em Proceedings of the IEEE/CVF International Conference on
  Computer Vision}, pages 16383--16392, 2021.

\bibitem{han2022divide}
Xianjing Han, Xingning Dong, Xuemeng Song, Tian Gan, Yibing Zhan, Yan Yan, and
  Liqiang Nie.
\newblock Divide-and-conquer predictor for unbiased scene graph generation.
\newblock {\em IEEE Transactions on Circuits and Systems for Video Technology},
  2022.

\bibitem{hung2020contextual}
Zih-Siou Hung, Arun Mallya, and Svetlana Lazebnik.
\newblock Contextual translation embedding for visual relationship detection
  and scene graph generation.
\newblock {\em IEEE transactions on pattern analysis and machine intelligence},
  43(11):3820--3832, 2020.

\bibitem{jacobs1991adaptive}
Robert~A Jacobs, Michael~I Jordan, Steven~J Nowlan, and Geoffrey~E Hinton.
\newblock Adaptive mixtures of local experts.
\newblock {\em Neural computation}, 3(1):79--87, 1991.

\bibitem{kenfack2020robotvqa}
Franklin~Kenghagho Kenfack, Feroz~Ahmed Siddiky, Ferenc Balint-Benczedi, and
  Michael Beetz.
\newblock Robotvqa—a scene-graph-and deep-learning-based visual question
  answering system for robot manipulation.
\newblock In {\em 2020 IEEE/RSJ International Conference on Intelligent Robots
  and Systems (IROS)}, pages 9667--9674. IEEE, 2020.

\bibitem{khandelwal2021segmentation}
Siddhesh Khandelwal, Mohammed Suhail, and Leonid Sigal.
\newblock Segmentation-grounded scene graph generation.
\newblock In {\em Proceedings of the IEEE/CVF International Conference on
  Computer Vision}, 2021.

\bibitem{krishna2017visual}
Ranjay Krishna, Yuke Zhu, Oliver Groth, Justin Johnson, Kenji Hata, Joshua
  Kravitz, Stephanie Chen, Yannis Kalantidis, Li-Jia Li, David~A Shamma, et~al.
\newblock Visual genome: Connecting language and vision using crowdsourced
  dense image annotations.
\newblock {\em International Journal of Computer Vision}, 123(1):32--73, 2017.

\bibitem{li2021bipartite}
Rongjie Li, Songyang Zhang, Bo Wan, and Xuming He.
\newblock Bipartite graph network with adaptive message passing for unbiased
  scene graph generation.
\newblock In {\em Proceedings of the IEEE/CVF Conference on Computer Vision and
  Pattern Recognition}, pages 11109--11119, 2021.

\bibitem{li2022ppdl}
Wei Li, Haiwei Zhang, Qijie Bai, Guoqing Zhao, Ning Jiang, and Xiaojie Yuan.
\newblock Ppdl: Predicate probability distribution based loss for unbiased
  scene graph generation.
\newblock In {\em Proceedings of the IEEE/CVF Conference on Computer Vision and
  Pattern Recognition}, pages 19447--19456, 2022.

\bibitem{li2017scene}
Yikang Li, Wanli Ouyang, Bolei Zhou, Kun Wang, and Xiaogang Wang.
\newblock Scene graph generation from objects, phrases and region captions.
\newblock In {\em Proceedings of the IEEE International Conference on Computer
  Vision}, pages 1261--1270, 2017.

\bibitem{lin2017focal}
Tsung-Yi Lin, Priya Goyal, Ross Girshick, Kaiming He, and Piotr Doll{\'a}r.
\newblock Focal loss for dense object detection.
\newblock In {\em Proceedings of the IEEE international conference on computer
  vision}, pages 2980--2988, 2017.

\bibitem{lin2020gps}
Xin Lin, Changxing Ding, Jinquan Zeng, and Dacheng Tao.
\newblock Gps-net: Graph property sensing network for scene graph generation.
\newblock In {\em Proceedings of the IEEE/CVF Conference on Computer Vision and
  Pattern Recognition}, pages 3746--3753, 2020.

\bibitem{liu2019large}
Ziwei Liu, Zhongqi Miao, Xiaohang Zhan, Jiayun Wang, Boqing Gong, and Stella~X
  Yu.
\newblock Large-scale long-tailed recognition in an open world.
\newblock In {\em Proceedings of the IEEE/CVF Conference on Computer Vision and
  Pattern Recognition}, pages 2537--2546, 2019.

\bibitem{lu2021context}
Yichao Lu, Himanshu Rai, Jason Chang, Boris Knyazev, Guangwei Yu, Shashank
  Shekhar, Graham~W Taylor, and Maksims Volkovs.
\newblock Context-aware scene graph generation with seq2seq transformers.
\newblock In {\em Proceedings of the IEEE/CVF International Conference on
  Computer Vision}, pages 15931--15941, 2021.

\bibitem{peng2019unsupervised}
Yuxin Peng and Jingze Chi.
\newblock Unsupervised cross-media retrieval using domain adaptation with scene
  graph.
\newblock {\em IEEE Transactions on Circuits and Systems for Video Technology},
  30(11):4368--4379, 2019.

\bibitem{qi2019attentive}
Mengshi Qi, Weijian Li, Zhengyuan Yang, Yunhong Wang, and Jiebo Luo.
\newblock Attentive relational networks for mapping images to scene graphs.
\newblock In {\em Proceedings of the IEEE/CVF Conference on Computer Vision and
  Pattern Recognition}, pages 3957--3966, 2019.

\bibitem{ren2015faster}
Shaoqing Ren, Kaiming He, Ross Girshick, and Jian Sun.
\newblock Faster r-cnn: Towards real-time object detection with region proposal
  networks.
\newblock In {\em Advances in neural information processing systems}, pages
  91--99, 2015.

\bibitem{riquelme2021scaling}
Carlos Riquelme, Joan Puigcerver, Basil Mustafa, Maxim Neumann, Rodolphe
  Jenatton, Andr{\'e} Susano~Pinto, Daniel Keysers, and Neil Houlsby.
\newblock Scaling vision with sparse mixture of experts.
\newblock {\em Advances in Neural Information Processing Systems}, 34, 2021.

\bibitem{rosinol20203d}
Antoni Rosinol, Arjun Gupta, Marcus Abate, Jingnan Shi, and Luca Carlone.
\newblock 3d dynamic scene graphs: Actionable spatial perception with places,
  objects, and humans.
\newblock {\em Robotics: Science and Systems (RSS)}, 2020.

\bibitem{rosinol2021kimera}
Antoni Rosinol, Andrew Violette, Marcus Abate, Nathan Hughes, Yun Chang,
  Jingnan Shi, Arjun Gupta, and Luca Carlone.
\newblock Kimera: From slam to spatial perception with 3d dynamic scene graphs.
\newblock {\em The International Journal of Robotics Research},
  40(12-14):1510--1546, 2021.

\bibitem{suhail2021energy}
Mohammed Suhail, Abhay Mittal, Behjat Siddiquie, Chris Broaddus, Jayan Eledath,
  Gerard Medioni, and Leonid Sigal.
\newblock Energy-based learning for scene graph generation.
\newblock In {\em Proceedings of the IEEE/CVF Conference on Computer Vision and
  Pattern Recognition}, pages 13936--13945, 2021.

\bibitem{tang2020sggcode}
Kaihua Tang.
\newblock A scene graph generation codebase in pytorch, 2020.
\newblock \url{https://github.com/KaihuaTang/Scene-Graph-Benchmark.pytorch}.

\bibitem{tang2020unbiased}
Kaihua Tang, Yulei Niu, Jianqiang Huang, Jiaxin Shi, and Hanwang Zhang.
\newblock Unbiased scene graph generation from biased training.
\newblock In {\em Proceedings of the IEEE/CVF Conference on Computer Vision and
  Pattern Recognition}, pages 3716--3725, 2020.

\bibitem{tang2019learning}
Kaihua Tang, Hanwang Zhang, Baoyuan Wu, Wenhan Luo, and Wei Liu.
\newblock Learning to compose dynamic tree structures for visual contexts.
\newblock In {\em Proceedings of the IEEE/CVF Conference on Computer Vision and
  Pattern Recognition}, pages 6619--6628, 2019.

\bibitem{wang2019exploring}
Wenbin Wang, Ruiping Wang, Shiguang Shan, and Xilin Chen.
\newblock Exploring context and visual pattern of relationship for scene graph
  generation.
\newblock In {\em Proceedings of the IEEE/CVF Conference on Computer Vision and
  Pattern Recognition}, pages 8188--8197, 2019.

\bibitem{wang2020sketching}
Wenbin Wang, Ruiping Wang, Shiguang Shan, and Xilin Chen.
\newblock Sketching image gist: Human-mimetic hierarchical scene graph
  generation.
\newblock In {\em European Conference on Computer Vision}, pages 222--239.
  Springer, 2020.

\bibitem{wang2021longtailed}
Xudong Wang, Long Lian, Zhongqi Miao, Ziwei Liu, and Stella Yu.
\newblock Long-tailed recognition by routing diverse distribution-aware
  experts.
\newblock In {\em International Conference on Learning Representations}, 2021.

\bibitem{wang2019learning}
Xin Wang, Jiawei Wu, Da Zhang, Yu Su, and William~Yang Wang.
\newblock Learning to compose topic-aware mixture of experts for zero-shot
  video captioning.
\newblock In {\em Proceedings of the AAAI Conference on Artificial
  Intelligence}, volume~33, pages 8965--8972, 2019.

\bibitem{Xie2016}
Saining Xie, Ross Girshick, Piotr Dollár, Zhuowen Tu, and Kaiming He.
\newblock Aggregated residual transformations for deep neural networks.
\newblock {\em arXiv preprint arXiv:1611.05431}, 2016.

\bibitem{xu2017scene}
Danfei Xu, Yuke Zhu, Christopher~B Choy, and Li Fei-Fei.
\newblock Scene graph generation by iterative message passing.
\newblock In {\em Proceedings of the IEEE conference on computer vision and
  pattern recognition}, pages 5410--5419, 2017.

\bibitem{xu2020scene}
Ning Xu, An-An Liu, Yongkang Wong, Weizhi Nie, Yuting Su, and Mohan
  Kankanhalli.
\newblock Scene graph inference via multi-scale context modeling.
\newblock {\em IEEE Transactions on Circuits and Systems for Video Technology},
  31(3):1031--1041, 2020.

\bibitem{yan2020pcpl}
Shaotian Yan, Chen Shen, Zhongming Jin, Jianqiang Huang, Rongxin Jiang, Yaowu
  Chen, and Xian-Sheng Hua.
\newblock Pcpl: Predicate-correlation perception learning for unbiased scene
  graph generation.
\newblock In {\em Proceedings of the 28th ACM International Conference on
  Multimedia}, pages 265--273, 2020.

\bibitem{yang2021probabilistic}
Gengcong Yang, Jingyi Zhang, Yong Zhang, Baoyuan Wu, and Yujiu Yang.
\newblock Probabilistic modeling of semantic ambiguity for scene graph
  generation.
\newblock In {\em Proceedings of the IEEE/CVF Conference on Computer Vision and
  Pattern Recognition}, pages 12527--12536, 2021.

\bibitem{yang2018graph}
Jianwei Yang, Jiasen Lu, Stefan Lee, Dhruv Batra, and Devi Parikh.
\newblock Graph r-cnn for scene graph generation.
\newblock In {\em Proceedings of the European conference on computer vision
  (ECCV)}, pages 670--685, 2018.

\bibitem{yang2019auto}
Xu Yang, Kaihua Tang, Hanwang Zhang, and Jianfei Cai.
\newblock Auto-encoding scene graphs for image captioning.
\newblock In {\em Proceedings of the IEEE/CVF Conference on Computer Vision and
  Pattern Recognition}, pages 10685--10694, 2019.

\bibitem{yu2020cogtree}
Jing Yu, Yuan Chai, Yujing Wang, Yue Hu, and Qi Wu.
\newblock Cogtree: Cognition tree loss for unbiased scene graph generation.
\newblock {\em arXiv preprint arXiv:2009.07526}, 2020.

\bibitem{zareian2020bridging}
Alireza Zareian, Svebor Karaman, and Shih-Fu Chang.
\newblock Bridging knowledge graphs to generate scene graphs.
\newblock In {\em European Conference on Computer Vision}, pages 606--623.
  Springer, 2020.

\bibitem{zareian2020learning}
Alireza Zareian, Zhecan Wang, Haoxuan You, and Shih-Fu Chang.
\newblock Learning visual commonsense for robust scene graph generation.
\newblock In {\em European Conference on Computer Vision}, pages 642--657.
  Springer, 2020.

\bibitem{zellers2018neural}
Rowan Zellers, Mark Yatskar, Sam Thomson, and Yejin Choi.
\newblock Neural motifs: Scene graph parsing with global context.
\newblock In {\em Proceedings of the IEEE Conference on Computer Vision and
  Pattern Recognition}, pages 5831--5840, 2018.

\bibitem{zhang2019graphical}
Ji Zhang, Kevin~J Shih, Ahmed Elgammal, Andrew Tao, and Bryan Catanzaro.
\newblock Graphical contrastive losses for scene graph parsing.
\newblock In {\em Proceedings of the IEEE/CVF Conference on Computer Vision and
  Pattern Recognition}, pages 11535--11543, 2019.

\bibitem{zhang2022boosting}
Yong Zhang, Yingwei Pan, Ting Yao, Rui Huang, Tao Mei, and Chang-Wen Chen.
\newblock Boosting scene graph generation with visual relation saliency.
\newblock {\em ACM Transactions on Multimedia Computing, Communications, and
  Applications (TOMM)}, 2022.

\bibitem{zhao2021semantically}
Bowen Zhao, Zhendong Mao, Shancheng Fang, Wenyu Zang, and Yongdong Zhang.
\newblock Semantically similarity-wise dual-branch network for scene graph
  generation.
\newblock {\em IEEE Transactions on Circuits and Systems for Video Technology},
  2021.

\bibitem{zhao2020towards}
Shuai Zhao, Liguang Zhou, Wenxiao Wang, Deng Cai, Tin~Lun Lam, and Yangsheng
  Xu.
\newblock Towards better accuracy-efficiency trade-offs: Divide and
  co-training.
\newblock {\em IEEE Transactions on Image Processing}, 2022.

\bibitem{zhong2020comprehensive}
Yiwu Zhong, Liwei Wang, Jianshu Chen, Dong Yu, and Yin Li.
\newblock Comprehensive image captioning via scene graph decomposition.
\newblock In {\em European Conference on Computer Vision}, pages 211--229.
  Springer, 2020.

\bibitem{zhou2022debiased}
Hao Zhou, Jun Zhang, Tingjin Luo, Yazhou Yang, and Jun Lei.
\newblock Debiased scene graph generation for dual imbalance learning.
\newblock {\em IEEE Transactions on Pattern Analysis and Machine Intelligence},
  2022.

\bibitem{zhu2021hierarchical}
Yifeng Zhu, Jonathan Tremblay, Stan Birchfield, and Yuke Zhu.
\newblock Hierarchical planning for long-horizon manipulation with geometric
  and symbolic scene graphs.
\newblock In {\em 2021 IEEE International Conference on Robotics and Automation
  (ICRA)}, pages 6541--6548. IEEE, 2021.

\end{thebibliography}
}
	
\end{document}